\documentclass[10pt,twocolumn,letterpaper]{article}
\usepackage[pagenumbers]{cvpr}
\usepackage{multirow,placeins}
\definecolor{cvprblue}{rgb}{0.21,0.49,0.74}
\usepackage[breaklinks,colorlinks,allcolors=cvprblue]{hyperref}

\makeatletter
\patchcmd{\@maketitle}{\vspace*{24pt}}{\vspace*{9pt}}{}{}
\patchcmd{\@maketitle}{\vspace*{12pt}}{\vspace*{4pt}}{}{}
\makeatother
\title{When Semantics Matter: Reliability-Aware Semantic--Rhythm Control for Co-Speech Gesture Generation}
\author{
Zhirui Xing\textsuperscript{1\ensuremath{*}},
Ming Meng\textsuperscript{2\ensuremath{*\dagger}},
Kaige Li\textsuperscript{3},
Ziyi Xu\textsuperscript{1},
Zhaoxin Fan\textsuperscript{4\ensuremath{\dagger}}\\[2pt]
{\small\shortstack[c]{
  \textsuperscript{1}Hainan International College, Communication University of China, Lingshui, China\\
  \textsuperscript{2}School of Data Science and Intelligent Media, Communication University of China, Beijing, China\\
  \textsuperscript{3}School of Cyber Science and Technology, Sun Yat-sen University, Shenzhen, China\\
  \textsuperscript{4}School of Artificial Intelligence, Beihang University, Beijing, China}}\\[2pt]
{\footnotesize \textsuperscript{\ensuremath{*}}These authors contributed equally.\quad
  \textsuperscript{\ensuremath{\dagger}}Corresponding authors:
  \href{mailto:mengming@cuc.edu.cn}{mengming@cuc.edu.cn},
  \href{mailto:zhaoxinf@buaa.edu.cn}{zhaoxinf@buaa.edu.cn}.}
}
\hypersetup{
  pdftitle={When Semantics Matter: Reliability-Aware Semantic-Rhythm Control for Co-Speech Gesture Generation},
  pdfauthor={Zhirui Xing, Ming Meng, Kaige Li, Ziyi Xu, Zhaoxin Fan}
}

\begin{document}
\maketitle

\noindent\begin{minipage}[t]{\columnwidth}
  \centerline{\large\bfseries Abstract}
  \vspace{8pt}
  {\itshape\noindent Co-speech gesture generation aims to synthesize natural gestures that are both temporally synchronized with speech and semantically consistent with the spoken content. Although recent methods can generate rhythmically plausible motions, they often rely heavily on acoustic prosody while underutilizing textual semantics, especially when semantic annotations are incomplete, noisy, or unavailable. Consequently, the generated gestures may follow speech rhythm while failing to express the intended semantics. To address this problem, we propose a reliability-aware semantic--rhythm control framework for co-speech gesture generation. We first learn a discrete motion prior that represents continuous gestures in a compact and structured motion-code space. We then introduce a dual-branch semantic contribution estimation mechanism consisting of a full multimodal branch and an audio-only branch. Their distributional discrepancy provides a model-based proxy for text-induced predictive influence, which is interpreted jointly with segment-level semantic relevance. Based on this estimate, a controllable semantic--rhythm objective selectively strengthens semantic guidance in content-relevant segments while limiting unnecessary semantic intervention in rhythm-dominant segments. Furthermore, we treat background noise as an acoustic reliability condition and introduce noise-conditioned feature modulation together with benefit-regularized latent perturbation to improve generation robustness under realistic acoustic environments. Experiments on benchmark datasets demonstrate that the proposed framework achieves a favorable balance among semantic expressiveness, rhythmic synchronization, motion diversity, and robustness, enabling reliable and controllable co-speech gesture generation.\par}
\end{minipage}\par

\begin{figure}[!t]
  \centering
  \includegraphics[width=\columnwidth]{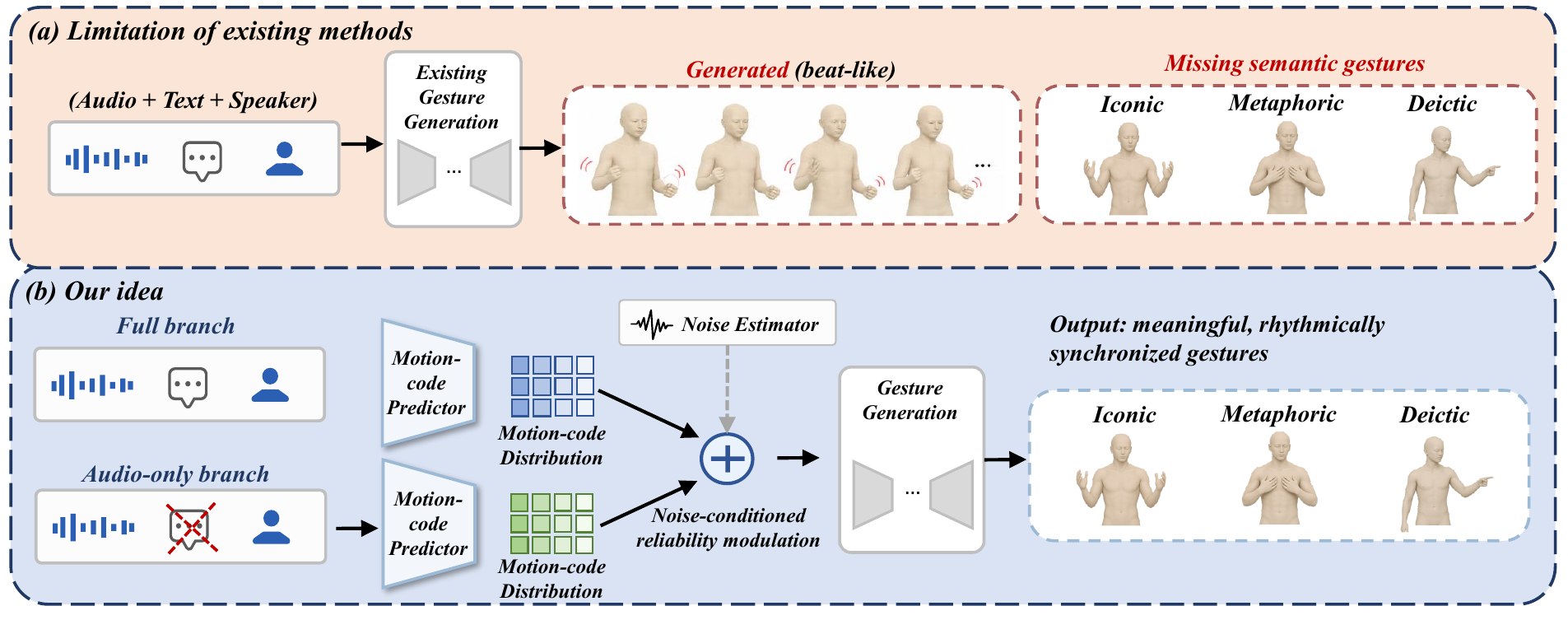}
  \caption{Motivation for segment-wise semantic guidance and acoustic reliability conditioning in co-speech gesture generation.}
  \label{motivation}
\end{figure}

\section{Introduction}
Co-speech gestures complement spoken language. They convey emphasis, discourse structure, affect, and semantic intent. Their automatic generation is therefore important for digital humans, embodied agents, virtual avatars, and multimodal human--computer interaction \citep{ref09,ref17,ref24}. An effective system should generate natural body and hand motions. These motions should be temporally synchronized with speech and semantically consistent with the utterance.

Data-driven approaches have substantially improved motion naturalness, diversity, and rhythmic synchronization \citep{ref03,ref19,ref21}. Early studies used recurrent, adversarial, and flow-based models to learn audio--gesture correlations and speaker-specific motion styles \citep{ref01,ref02,ref06,ref24}. Later work explored conditional variational autoencoders, hierarchical cross-modal learning, diffusion models, and vector-quantized motion priors \citep{ref08,ref14}. These methods can generate smooth motions that are well synchronized with speech. However, many rely mainly on low-level acoustic cues, such as pitch, energy, speaking rate, and beat patterns \citep{ref23,ref28}. These cues are useful for rhythmic gestures. They are less effective for semantically grounded iconic, metaphoric, and deictic gestures.

Textual conditioning introduces high-level semantics into gesture generation \citep{ref07}. Recent work uses language representations, semantic coherence, retrieval, motion examples, and gesture captions to improve grounding and control \citep{ref04,ref05,ref11,ref22,ref26,ref27}. Yet semantic cues are sparse: some phrases invite content-bearing gestures, while other segments primarily follow prosody. Uniform fusion can introduce irrelevant textual cues or weaken rhythmic alignment \citep{ref10,ref15}. Moreover, text can be present without changing a generator's predictions, especially when acoustic features dominate. Background noise further corrupts these features and can disrupt alignment; streaming studies also highlight the difficulty of imperfect speech input \citep{ref12,ref20}. These observations motivate explicit modeling of semantic contribution and acoustic reliability.

As illustrated in Fig.~\ref{motivation}, we propose a reliability-aware semantic--rhythm framework. A discrete prior first maps continuous gestures to motion codes. We then compare full (audio, text, and speaker) and audio-only predictions in this shared code space. Their distributional discrepancy measures text-induced predictive changes, while a separate relevance estimator indicates where those changes are desirable. A segment-wise objective encourages textual influence in content-relevant segments and limits it in rhythm-dominant segments. Finally, acoustic-condition modulation and benefit-regularized latent perturbation address corrupted speech input.

The main contributions are summarized as follows:
\begin{itemize}
  \item We propose a reliability-aware semantic contribution modeling framework that combines a discrete motion prior with multimodal conditioning to generate stable, semantically meaningful, and rhythmically synchronized co-speech gestures.
  \item We introduce a dual-branch mechanism that measures text-induced predictive changes through a conditional information-gain proxy and combines this signal with semantic relevance to guide a controllable semantic--rhythm optimization objective.
  \item We develop an acoustic reliability modulation strategy with benefit-regularized latent perturbation, enabling the generator to adapt to background noise and improve robustness under realistic acoustic conditions.
\end{itemize}

\begin{figure*}[!t]
  \centering
  \includegraphics[width=\textwidth]{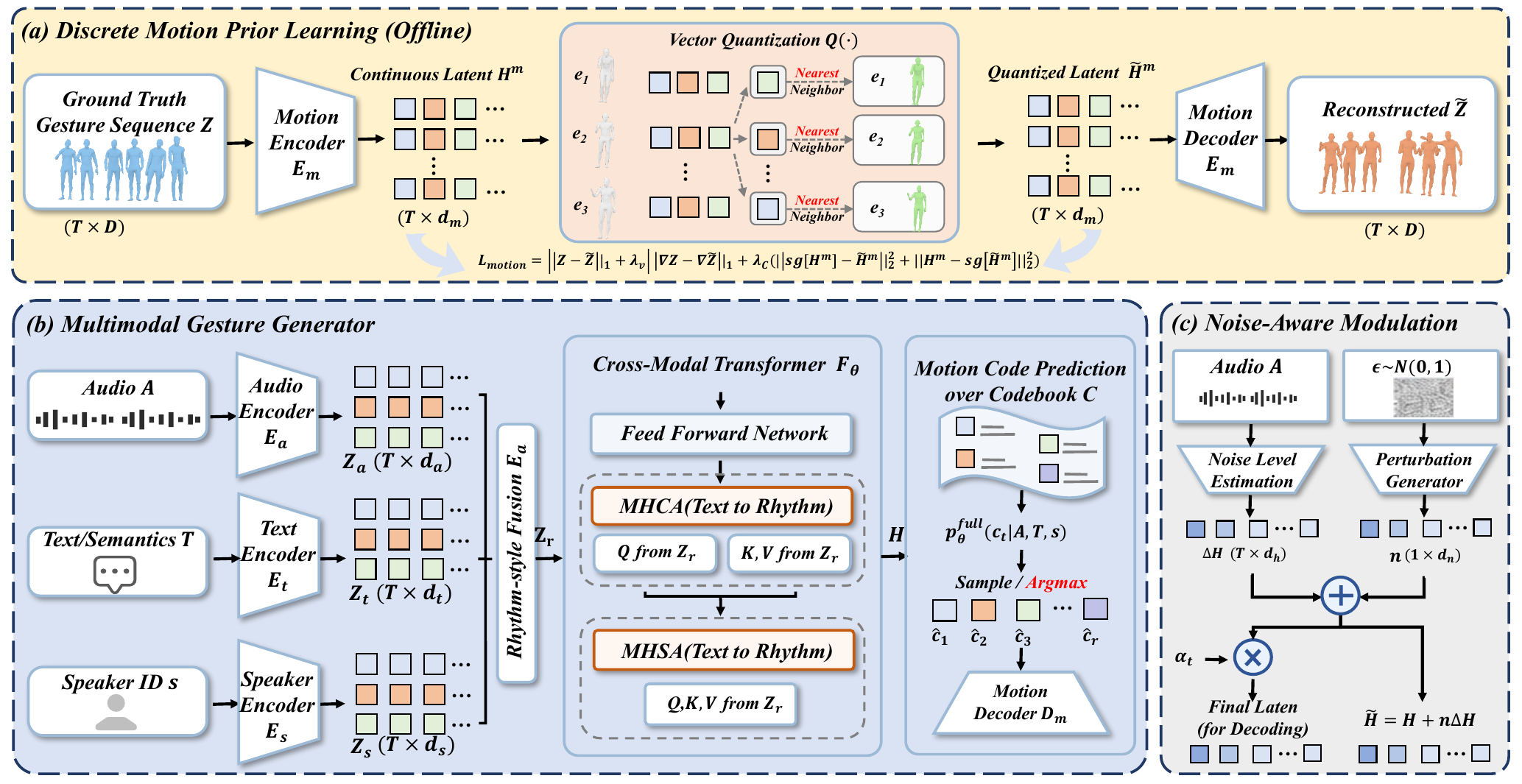}
  \caption{Overview of the proposed reliability-aware semantic contribution modeling framework.}
  \label{framework}
\end{figure*}

\section{Related Work}
\subsection{Speech-Driven Gesture Generation}
Speech-driven gesture generation aims to synthesize human motions that are natural, diverse, and temporally aligned with speech \citep{ref01,ref06}. Early data-driven methods employed recurrent and adversarial architectures to jointly model acoustic, textual, and speaker-specific conditions \citep{ref24}. Normalizing-flow models subsequently improved motion diversity and enabled explicit control over gesture styles \citep{ref02}, while conditional variational autoencoders learned one-to-many mappings from speech audio to plausible gesture sequences \citep{ref08,ref19}. Recent approaches increasingly rely on attention-based and generative architectures. Hierarchical cross-modal modeling has been introduced to capture associations between speech and motion at multiple temporal levels \citep{ref03,ref13}. Diffusion-based methods improve motion quality by learning iterative denoising processes \citep{ref28}, whereas masked modeling and discrete motion representations support unified full-body generation \citep{ref10,ref21}. More recent studies explore flow matching with spatial--temporal body-region modeling \citep{ref12} and causal motion tokenization for streamable generation \citep{ref20}. Although these methods substantially improve realism, diversity, and rhythmic alignment, they generally do not explicitly quantify how much textual semantics contributes to each gesture prediction.

\subsection{Semantic-Aware Gesture Generation}
Textual information has been incorporated into co-speech gesture generation to improve semantic consistency \citep{ref07,ref26}. LivelySpeaker retrieves semantically relevant motion units and integrates them with a diffusion-based generator \citep{ref27}. SemGes models semantic coherence and local semantic relevance to improve the grounding of generated gestures in spoken content \citep{ref11}. GestureHYDRA further combines a hybrid-modality diffusion Transformer with retrieval-augmented generation to activate semantically explicit hand gestures \citep{ref22}. Beyond transcript conditioning, recent methods introduce richer semantic control signals. MECo uses large language models to interpret motion examples and speech conditions, enabling example-guided and body-part-level control \citep{ref04}. CoordSpeaker constructs gesture captions and employs them as semantic conditions for coordinated gesture synthesis \citep{ref05}. These methods demonstrate the value of language and high-level semantic priors. However, most of them focus on strengthening or controlling semantic conditioning under predefined semantic inputs. They do not explicitly measure whether textual information provides additional predictive utility beyond acoustic prosody, nor do they distinguish semantic-dominant segments from rhythm-dominant ones.

\section{Method}
\subsection{Problem Formulation and Framework Overview}
Given an acoustic sequence $A$, its textual-semantic representation $T$, and a speaker identity $s$, co-speech gesture generation aims to synthesize a gesture sequence $Z$ that is temporally synchronized with speech, semantically consistent with the spoken content, and compatible with the speaker-specific motion style. Accordingly, the model learns the conditional distribution $p_\theta(Z\mid A,T,s)$.

As shown in Fig.~\ref{framework}, our framework regulates textual influence at the segment level while accounting for the reliability of the acoustic input. It consists of four components: a discrete motion prior, a dual-branch multimodal motion-code generator, conditional semantic information gain with controllable semantic--rhythm optimization, and acoustic reliability conditioning with benefit-regularized latent perturbation.

\subsection{Discrete Motion Prior Learning}
Direct regression of high-dimensional joint trajectories frequently produces jittering motion and unstable temporal transitions. We therefore learn a discrete motion prior that maps continuous gesture sequences into a compact and reusable motion-code space. This representation allows the subsequent multimodal generator to predict structured motion tokens rather than unconstrained frame-level joint coordinates.

Let $Z=\{z_t\}_{t=1}^{L}$ denote a gesture sequence, where $z_t\in\mathbb{R}^{J\times d}$ represents the pose at frame $t$, $J$ is the number of joints, and $d$ is the coordinate dimensionality. We divide the sequence into $N=\lceil L/K\rceil$ temporal segments of $K$ frames. A motion encoder $E_m$ transforms these segments into latent representations $H^m=\{h_i^m\}_{i=1}^{N}$, where $h_i^m\in\mathbb{R}^{D_m}$. The encoder combines temporal convolutions, which capture short-range motion dynamics, with Transformer blocks that model dependencies across motion segments.

A learnable codebook $\mathcal{E}=\{e_k\}_{k=1}^{M}$ contains $M$ motion prototypes. Each segment representation is assigned to its nearest codebook entry:
\begin{equation}
  q_i=\arg\min_{k\in\{1,\ldots,M\}}\left\|h_i^m-e_k\right\|_2^2,
  \qquad \bar{h}_i^m=e_{q_i}.
\end{equation}

The resulting index sequence $C^*=\{q_i\}_{i=1}^{N}$ provides a symbolic representation of the target gesture. A motion decoder $D_m$ reconstructs the continuous sequence from the quantized representations $\bar{H}^m=\{\bar{h}_i^m\}_{i=1}^{N}$. It employs Transformer blocks followed by temporal upsampling layers to recover both long-range motion dependencies and frame-level pose trajectories.

The motion prior is trained with pose reconstruction, velocity consistency, codebook updating, and commitment regularization:
\begin{equation}
  \begin{aligned}
    \mathcal{L}_{\mathrm{prior}}={}&
    \left\|Z-\widetilde{Z}\right\|_1
    +\lambda_{\mathrm{vel}}\left\|\nabla Z-\nabla\widetilde{Z}\right\|_1\\
    &+\lambda_{\mathrm{cb}}\left\|\operatorname{sg}[H^m]-\bar{H}^m\right\|_2^2\\
    &+\lambda_{\mathrm{com}}\left\|H^m-\operatorname{sg}[\bar{H}^m]\right\|_2^2.
  \end{aligned}
\end{equation}
where $\widetilde{Z}=D_m(\bar{H}^m)$, $\nabla$ denotes the temporal velocity operator, and $\operatorname{sg}[\cdot]$ is the stop-gradient operation. The first two terms preserve pose accuracy and temporal continuity, whereas the last two stabilize codebook learning and encourage encoder outputs to remain close to the selected motion prototypes.

After this stage, the learned codebook defines the discrete prediction space for multimodal generation. The motion decoder is retained to reconstruct continuous gestures from the predicted codes.

\subsection{Dual-Branch Multimodal Motion-Code Generation}
The multimodal generator predicts the target code sequence $C^*$ from speech, text, and speaker identity. Acoustic features are encoded by an audio encoder, while the speaker identity is mapped to a style embedding. Their representations are fused into a rhythm--style stream that captures prosodic timing and speaker-dependent motion characteristics. In parallel, a text encoder extracts contextual semantic representations from the transcript.

\begin{figure}[!t]
  \centering
  \includegraphics[width=\columnwidth]{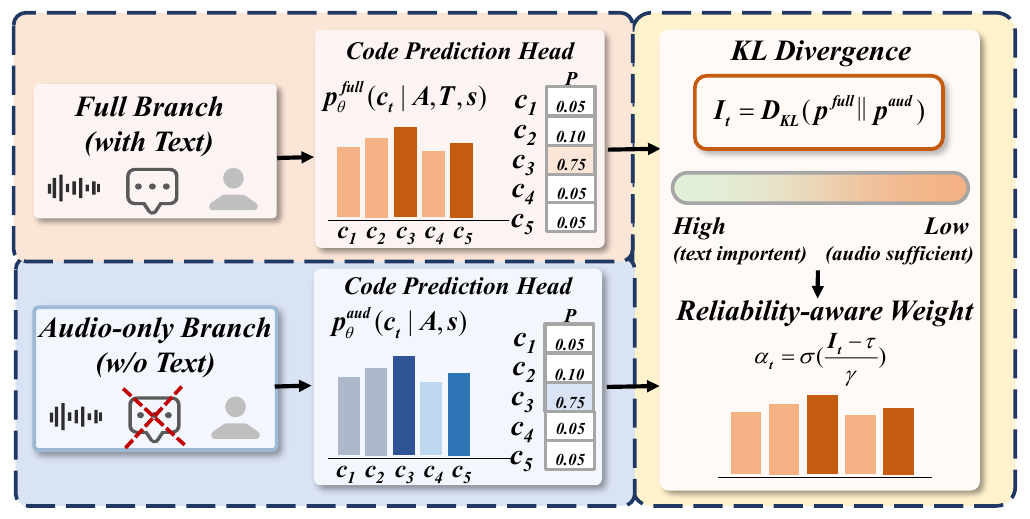}
  \caption{Dual-branch semantic contribution estimation. Both branches predict distributions over the same motion codebook. KL divergence measures text-induced predictive changes, while a separate relevance estimator guides the semantic--rhythm objective. Stop-gradient is applied to the audio-only reference and relevance weights within this objective.}
  \label{dual}
\end{figure}

As shown in Fig.~\ref{dual}, we construct two predictive branches. The full branch uses a cross-modal Transformer: rhythm--style features query the contextual text features, which provide the keys and values. The audio-only branch follows the same rhythm modeling pathway but does not receive textual information. Both branches predict categorical distributions over the same frozen motion codebook:
\begin{equation}
  p_i^{\mathrm{full}}=p_\theta(c_i\mid A,T,s),\quad
  p_i^{\mathrm{aud}}=p_{\theta_a}(c_i\mid A,s),
\end{equation}
where $i$ indexes motion segments. The two branches share the rhythm encoder and motion-code prediction space, providing comparable predictions in a common representation. Their discrepancy measures changes in the predicted code distribution associated with adding text; it is not an independent causal estimate of semantic correctness.

Both distributions are supervised using the ground-truth motion codes obtained from the pretrained motion prior:
\begin{equation}
  \mathcal{L}_{\mathrm{code}}=-\frac{1}{N}\sum_{i=1}^{N}
  \left[\log p_i^{\mathrm{full}}(q_i)+\lambda_{\mathrm{aud}}\log p_i^{\mathrm{aud}}(q_i)\right].
\end{equation}

During training, the full-branch distribution can be converted into a differentiable soft motion representation by taking the probability-weighted combination of codebook embeddings. The frozen motion decoder then reconstructs a continuous gesture sequence from these soft embeddings, enabling gesture-level reconstruction supervision. During inference, hard motion codes are obtained by categorical sampling or maximum-probability selection and decoded by $D_m$.

We additionally encourage global text--motion coherence by projecting pooled textual and generated motion representations into a shared embedding space and minimizing their cosine distance. This auxiliary objective complements code-level supervision by discouraging semantically inconsistent gesture sequences.

\subsection{Conditional Semantic Information Gain and Semantic--Rhythm Control}
Merely providing textual features to a multimodal generator does not ensure that the model actually uses them. A sufficiently strong acoustic pathway may still dominate the prediction, particularly when semantic cues are weak, missing, or noisy. We therefore quantify the effect of textual information through the discrepancy between the full and audio-only predictive distributions.

For each motion segment, we define a conditional semantic information-gain (CIG) proxy as
\begin{equation}
  \mathcal{I}_i=D_{\mathrm{KL}}\!\left(p_i^{\mathrm{full}}\mathbin{\|}\operatorname{sg}[p_i^{\mathrm{aud}}]\right).
\end{equation}
A large $\mathcal{I}_i$ indicates a substantial text-induced change in the predicted motion-code distribution; it does not by itself establish semantic correctness. A small value indicates similar predictions with and without text. We therefore interpret this discrepancy jointly with semantic relevance and motion supervision, rather than as a direct estimate of semantic utility or conditional mutual information.

Because semantic importance varies over time, a lightweight estimator predicts a relevance score $\rho_i\in[0,1]$ from aligned acoustic, textual, and full-branch representations. The relevance estimator is trained with segment-level relevance targets through $\mathcal{L}_\rho$. The relevance score specifies where semantic influence is desired, whereas $\mathcal{I}_i$ measures how strongly text changes the prediction.

We regulate semantic influence with detached relevance weights $\bar{\rho}_i=\operatorname{sg}[\rho_i]$:
\begin{equation}
  \begin{aligned}
    \mathcal{L}_{\mathrm{IG}}={}&\frac{1}{N}\sum_{i=1}^{N}
    \Bigl(\lambda_{\mathrm{on}}\bar{\rho}_i[\tau-\mathcal{I}_i]_+\\
    &\qquad+\lambda_{\mathrm{off}}(1-\bar{\rho}_i)[\mathcal{I}_i-\zeta]_+\Bigr).
  \end{aligned}
\end{equation}
Here, $[x]_+=\max(0,x)$, $\tau$ is the semantic activation margin, and $\zeta$ is the rhythm-preservation margin, with $0\leq\zeta<\tau$. The first term encourages text-dependent predictions in semantically relevant segments; the second discourages unnecessary deviations from the audio-only reference in rhythm-dominant segments. Code-level and gesture-level supervision constrain these changes toward the target motion.

The audio-only distribution and relevance weights are detached only when computing this control objective. The audio-only branch remains supervised by its code prediction loss, and the relevance estimator remains supervised by its relevance targets. This prevents direct updates to the reference output or relevance scores merely to satisfy the discrepancy margins. Shared encoder parameters can still change through the other gradient paths. The margins act during training; beat alignment remains evaluated separately with BC.

\subsection{Acoustic Reliability Conditioning and Benefit-Regularized Latent Perturbation}
Real-world speech often contains stationary or transient background noise that corrupts prosodic features and weakens audio--gesture alignment. Rather than treating noise solely as an unwanted disturbance, we explicitly encode it as an acoustic reliability condition and use this condition to adapt the multimodal generator.

For each segment-aligned acoustic window, frames within the lowest energy quantile are treated as candidate background-dominant observations rather than noise-only samples. We summarize their spectral centroid, spectral flatness, spectral roll-off, and the mean and variance of the 64-dimensional log-mel representation to obtain a background-condition descriptor $B_i$. The descriptor is projected into a compact reliability embedding $z_i^n$. The descriptor captures background conditions; it is not an SNR estimate.

The reliability embedding modulates the hidden representations through noise-conditioned adaptive layer normalization:
\begin{equation}
  \begin{aligned}
    \operatorname{AdaLN}(h_i;z_i^n)
    &=(\gamma+\Delta\gamma_i)\odot\frac{h_i-\mu_i}{\sigma_i+\epsilon}\\
    &\quad+(\beta+\Delta\beta_i),
  \end{aligned}
\end{equation}
\begin{equation}
  [\Delta\gamma_i,\Delta\beta_i]=\operatorname{MLP}(z_i^n).
\end{equation}
where $\mu_i$ and $\sigma_i$ are the feature statistics, and $\epsilon$ is a numerical stability constant. This operation allows the generator to adjust its latent dynamics according to the estimated acoustic reliability rather than processing clean and noisy speech identically.

Beyond deterministic modulation, we introduce a benefit-regularized latent perturbation module that learns condition-dependent stochastic variations. Given the background condition $B_i$, speech-prosody feature $A_i^{\mathrm{sp}}$, textual feature $T_i$, and speaker identity $s$, the perturbation generator predicts a diagonal Gaussian distribution:
\begin{equation}
  \begin{aligned}
    &q_\phi(\varepsilon_i\mid B_i,A_i^{\mathrm{sp}},T_i,s)\\
    &\qquad=\mathcal{N}\!\left(0,\operatorname{diag}\!\left(
    \sigma_\phi^2(B_i,A_i^{\mathrm{sp}},T_i,s)\right)\right).
  \end{aligned}
\end{equation}
\begin{equation}
  \begin{aligned}
    \varepsilon_i&=\sigma_\phi(B_i,A_i^{\mathrm{sp}},T_i,s)\odot\xi_i,\\
    \xi_i&\sim\mathcal{N}(0,I).
  \end{aligned}
\end{equation}
\begin{equation}
  h_i'=h_i+g_iW_\varepsilon\varepsilon_i.
\end{equation}
Here, $h_i$ denotes the latent representation after reliability-conditioned normalization, $W_\varepsilon$ projects the sampled perturbation into this latent space, and $g_i\in[0,1]$ is a gate predicted from $z_i^n$. The perturbed representation $h_i'$ is passed to the motion-code prediction head. The gate adapts perturbation strength to the observed condition without enforcing monotonicity in the noise level.

Random latent noise does not necessarily improve generation quality. We therefore regularize the perturbation distribution with a reconstruction-ranking objective using a detached unperturbed reference:
\begin{equation}
  \mathcal{L}_{\mathrm{ben}}=\frac{1}{N}\sum_{i=1}^{N}
  \left[\ell_{\mathrm{rec}}^i(h_i')-\operatorname{sg}[\ell_{\mathrm{rec}}^i(h_i)]+m\right]_+.
\end{equation}
Here, $\ell_{\mathrm{rec}}^i(h_i')$ and $\ell_{\mathrm{rec}}^i(h_i)$ are segment-level pose and velocity reconstruction losses obtained by decoding the corresponding soft motion-code representations. Detaching the unperturbed loss prevents this objective from directly increasing the reference error. With $m=0$, the loss penalizes sampled perturbations only when their reconstruction error exceeds the detached unperturbed reference. This training constraint does not filter inference-time samples or guarantee an improvement for each draw.

A magnitude regularizer limits the size of the injected residual:
\begin{equation}
  \mathcal{L}_{\mathrm{mag}}=\frac{1}{N}\sum_{i=1}^{N}
  \left\|g_iW_\varepsilon\varepsilon_i\right\|_2^2.
\end{equation}
Together, these losses encourage restrained latent perturbations while retaining the code-level and reconstruction objectives as the main sources of motion supervision.

\begin{table*}[!t]
  \centering
  \caption{Comparison with representative co-speech gesture generation methods on BEAT~\citep{ref09} and TED-Expressive~\citep{ref14}.}
  \label{comparison}
  \small
  \begin{tabular}{lrrrrrrr}
    \toprule
    \multirow{2}{*}{Method} & \multicolumn{4}{c}{BEAT} & \multicolumn{3}{c}{TED-Expressive}\\
    \cmidrule(lr){2-5}\cmidrule(lr){6-8}
      & FGD$\downarrow$ & BC$\uparrow$ & Diversity$\uparrow$ & SRGR$\uparrow$
      & FGD$\downarrow$ & BC$\uparrow$ & Diversity$\uparrow$\\
    \midrule
    CaMN~\citep{ref09}       & 8.510 & 0.797 & 206.789 & 0.231 & 9.284 & 0.681 & 117.362\\
    SemGes~\citep{ref11}     & 4.762 & 0.453 & 305.706 & 0.256 & 5.106 & 0.713 & 139.824\\
    CoCoGesture~\citep{ref18}& 4.685 & 0.743 & 317.030 & 0.252 & 5.347 & 0.721 & 143.517\\
    EMAGE~\citep{ref10}      & 4.969 & 0.729 & 269.520 & 0.256 & 5.582 & 0.734 & 132.406\\
    GestureLSM~\citep{ref12} & 4.521 & 0.789 & 320.846 & 0.259 & 4.526 & 0.762 & 150.684\\
    LiveGesture~\citep{ref20}& 4.593 & 0.801 & 319.731 & 0.258 & 4.692 & 0.759 & 148.915\\
    GlobalDiff~\citep{ref25} & 4.558 & 0.795 & 318.426 & 0.257 & 4.611 & 0.754 & 149.638\\
    MIBURI~\citep{ref16}     & 4.447 & 0.803 & 321.734 & 0.260 & 4.406 & 0.769 & 152.316\\
    \midrule
    \textbf{Ours} & \textbf{4.362} & \textbf{0.805} & \textbf{322.455} & \textbf{0.261}
                  & \textbf{4.318} & \textbf{0.775} & \textbf{153.427}\\
    \bottomrule
  \end{tabular}
\end{table*}

\subsection{Training and Inference}
Training proceeds in two stages. First, the motion encoder, codebook, and decoder are optimized with $\mathcal{L}_{\mathrm{prior}}$. Second, the motion prior is frozen and the multimodal generator is trained using the target motion codes and decoded soft predictions. The second-stage objective is
\begin{equation}
  \begin{aligned}
    \mathcal{L}_{\mathrm{gen}}={}&\mathcal{L}_{\mathrm{code}}
    +\lambda_{\mathrm{rec}}\mathcal{L}_{\mathrm{rec}}+\lambda_{\mathrm{coh}}\mathcal{L}_{\mathrm{coh}}\\
    &+\lambda_\rho\mathcal{L}_\rho+\lambda_{\mathrm{IG}}\mathcal{L}_{\mathrm{IG}}\\
    &+\lambda_{\mathrm{ben}}\mathcal{L}_{\mathrm{ben}}+\lambda_{\mathrm{mag}}\mathcal{L}_{\mathrm{mag}}.
  \end{aligned}
\end{equation}
Here, $\mathcal{L}_{\mathrm{rec}}$ supervises pose and temporal-velocity reconstruction, $\mathcal{L}_{\mathrm{coh}}$ is the global text--motion cosine-distance objective, and $\mathcal{L}_\rho$ supervises relevance prediction. The remaining terms control text-induced predictive changes and regularize latent perturbations. Although the motion decoder is frozen, gradients through its soft inputs are retained to train the generator.

At inference, the full branch predicts hard motion codes, which the pretrained decoder converts into continuous gestures. The acoustic-condition embedding and perturbation gate are computed from the input speech. The learned distribution supplies the latent perturbation, and the resulting code predictions are decoded into poses. The audio-only reference is not used to select outputs at inference. The audio-only branch and relevance estimator serve the training objective and are omitted from standard generation. No ground-truth reconstruction loss is used to select inference outputs.

\section{Experiments}
\subsection{Datasets and Implementation Details}
We conduct experiments on BEAT \citep{ref09} and TED-Expressive \citep{ref14}. BEAT contains synchronized motion-capture, speech, text, and speaker annotations, whereas TED-Expressive provides in-the-wild upper-body gestures reconstructed from TED Talk videos. Following standard protocols, the two datasets are split into training, validation, and test sets with ratios of 19:2:2 and 8:1:1, respectively. Motion sequences are segmented into 34-frame clips, root-normalized, and aligned with the corresponding audio and text. The model is implemented in PyTorch and trained on a single NVIDIA GPU for 300 epochs using AdamW with a batch size of 64, an initial learning rate of $2 \times 10^{-4}$, and a weight decay of $1 \times 10^{-4}$. We adopt linear warm-up followed by cosine annealing and apply gradient clipping with a maximum norm of 1.0. Audio is resampled to 16 kHz, text features are extracted using a pretrained language encoder, and learnable embeddings represent speaker identities. Unless otherwise specified, the semantic contribution threshold is set to $\tau = 0.6$. We evaluate generation quality using Fr\'echet Gesture Distance (FGD), Beat Consistency (BC), Diversity, and Semantic-Relevant Gesture Recall (SRGR). Lower FGD indicates better motion realism, while higher BC, Diversity, and SRGR indicate stronger rhythmic alignment, motion variation, and semantic relevance, respectively.

\subsection{Comparison with State-of-the-Art Methods}
Table~\ref{comparison} compares the proposed method with representative co-speech gesture generation approaches on the BEAT and TED-Expressive datasets. Overall, our method achieves the best results across all reported metrics, indicating a favorable balance among motion realism, speech--gesture synchronization, semantic relevance, and motion diversity. On BEAT, our method obtains an FGD of 4.362, a BC score of 0.805, a Diversity score of 322.455, and an SRGR score of 0.261. Compared with the second-best MIBURI, it reduces FGD by 0.085 and improves BC, Diversity, and SRGR by 0.002, 0.721, and 0.001, respectively. The lower FGD indicates that the generated motions better approximate the real gesture distribution, while the improvements in BC and SRGR demonstrate stronger rhythmic synchronization and semantic consistency with the input speech. Meanwhile, the highest Diversity score shows that these gains are achieved without sacrificing motion variability. On TED-Expressive, our method also ranks first on all three metrics, achieving an FGD of 4.318, a BC score of 0.775, and a Diversity score of 153.427. Relative to MIBURI, it reduces FGD by 0.088 and improves BC and Diversity by 0.006 and 1.111, respectively. The same ordering on TED-Expressive extends the comparison to an in-the-wild corpus.

\begin{figure*}[!t]
  \centering
  \includegraphics[width=\textwidth]{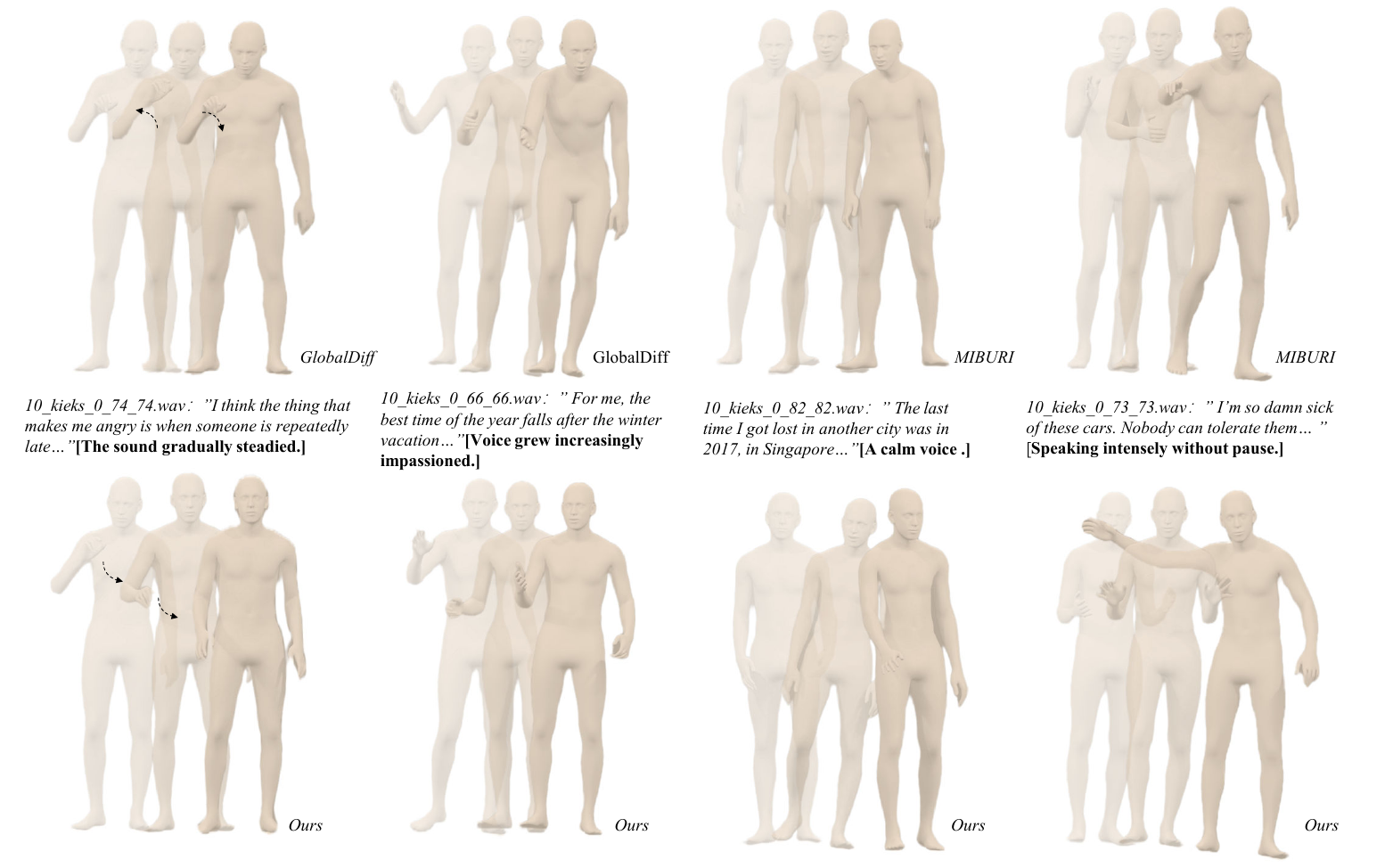}
  \caption{Qualitative comparison with GlobalDiff and MIBURI under gradually calming, increasingly impassioned, consistently calm, and continuously intense speech. Overlaid poses show temporal progression, and dashed arrows mark hand trajectories. Our method better follows changes in vocal intensity.}
  \label{qualitative}
\end{figure*}

\subsection{Qualitative Comparison}
Fig.~\ref{qualitative} presents qualitative comparisons under four representative speaking dynamics: gradually calming, increasingly impassioned, consistently calm, and continuously intense. Three temporally ordered poses are overlaid in each example to illustrate the evolution of the generated gesture, while the dashed arrows indicate the main hand trajectories. For gradually calming speech, GlobalDiff still produces relatively pronounced hand motion across the sequence and does not fully reflect the decreasing vocal intensity. In contrast, our method progressively reduces the motion amplitude and guides the hands toward a more stable pose, yielding a gesture transition consistent with the calming speech pattern.

For increasingly impassioned speech, our method produces a clearer expansion of the upper-body motion and stronger hand emphasis, whereas GlobalDiff exhibits a less distinctive change in gesture intensity. For consistently calm speech, MIBURI generates noticeable body and arm variations that are not fully aligned with the low-arousal speaking style. Our result maintains a restrained gesture range and a more stable body posture. Under continuously intense speech, both methods generate large-scale movements; however, MIBURI shows relatively dispersed pose transitions, while our method maintains sustained gesture intensity with a more coherent directional trajectory. These examples demonstrate that the proposed framework can adapt gesture magnitude, direction, and temporal evolution to different acoustic dynamics, rather than merely generating locally plausible poses.

\subsection{User Study}
As shown in Table~\ref{users}, our method received the highest mean rating on all three dimensions. Compared with MIBURI, the highest-rated baseline, our method scored 0.24 points higher in motion naturalness, 0.24 points higher in rhythmic alignment, and 0.26 points higher in semantic appropriateness. The largest mean difference is in semantic appropriateness, suggesting that participants perceived the generated gestures as more consistent with the meaning of the speech.

\begin{table}[!t]
  \centering
  \caption{User ratings of motion naturalness, rhythmic alignment, and semantic appropriateness. Values are mean scores on a 1--5 scale (higher is better), with 95\% confidence intervals in brackets.}
  \label{users}
  \resizebox{\columnwidth}{!}{%
    \begin{tabular}{@{}lccc@{}}
      \toprule
      Method & \shortstack{Motion\\naturalness$\uparrow$}
             & \shortstack{Rhythmic\\alignment$\uparrow$}
             & \shortstack{Semantic\\appropriateness$\uparrow$}\\
      \midrule
      GlobalDiff & 3.42 [3.26, 3.58] & 3.35 [3.19, 3.51] & 3.47 [3.31, 3.63]\\
      MIBURI     & 3.74 [3.57, 3.91] & 3.68 [3.51, 3.85] & 3.79 [3.62, 3.96]\\
      \textbf{Ours} & \textbf{3.98} [3.81, 4.15] & \textbf{3.92} [3.75, 4.09] & \textbf{4.05} [3.88, 4.22]\\
      \bottomrule
    \end{tabular}}
\end{table}

\begin{table*}[!t]
  \centering
  \caption{Semantic--rhythm control on BEAT. Overall metrics use the full test set; masking scores compare paired original and masked transcripts. Preference is the percentage of blind comparisons favoring the original transcript for semantic appropriateness. Dashes denote unreported results.}
  \label{semantics}
  \resizebox{\textwidth}{!}{%
    \begin{tabular}{@{}lrrrrccc@{}}
      \toprule
       & \multicolumn{4}{c}{Overall test performance} & \multicolumn{3}{c}{Transcript masking}\\
      \cmidrule(lr){2-5}\cmidrule(lr){6-8}
      Configuration & FGD$\downarrow$ & BC$\uparrow$ & Diversity$\uparrow$ & SRGR$\uparrow$
       & \shortstack{Semantic SRGR\\Original / Masked} & \shortstack{Beat BC\\Original / Masked}
       & \shortstack{Semantic preference\\for original (\%)}\\
      \midrule
      Direct Fusion & 5.084 & 0.789 & 314.672 & 0.241 & --- & --- & ---\\
      Dual Branch & 4.931 & 0.793 & 317.486 & 0.248 & --- & --- & ---\\
      Dual Branch + CIG ($\lambda_{\mathrm{IG}}=0$) & 4.846 & 0.799 & 320.318 & 0.254 & --- & --- & ---\\
      w/o $\mathcal{L}_{\mathrm{IG}}$ & 4.846 & 0.799 & 320.318 & 0.254 & 0.251 / 0.232 & 0.796 / 0.781 & 62\\
      Full Model & \textbf{4.362} & \textbf{0.805} & \textbf{322.455} & \textbf{0.261} & 0.260 / 0.227 & 0.803 / 0.776 & 71\\
      \bottomrule
    \end{tabular}}
\end{table*}

\begin{table*}[!t]
  \centering
  \caption{Ablation study of acoustic reliability modeling on BEAT.}
  \label{noise}
  \small
  \begin{tabular}{lrrrrrrrrr}
    \toprule
      & \multicolumn{3}{c}{Clean} & \multicolumn{3}{c}{10 dB} & \multicolumn{3}{c}{0 dB}\\
    \cmidrule(lr){2-4}\cmidrule(lr){5-7}\cmidrule(lr){8-10}
    Configuration & FGD$\downarrow$ & BC$\uparrow$ & SRGR$\uparrow$
                  & FGD$\downarrow$ & BC$\uparrow$ & SRGR$\uparrow$
                  & FGD$\downarrow$ & BC$\uparrow$ & SRGR$\uparrow$\\
    \midrule
    w/o Reliability  & 4.879 & 0.800 & 0.254 & 5.421 & 0.748 & 0.229 & 6.736 & 0.672 & 0.194\\
    AdaLN Only       & 4.824 & 0.802 & 0.256 & 5.183 & 0.769 & 0.241 & 6.081 & 0.706 & 0.211\\
    Perturbation Only& 4.837 & 0.801 & 0.255 & 5.246 & 0.763 & 0.238 & 6.204 & 0.699 & 0.207\\
    Full Model       & \textbf{4.362} & \textbf{0.805} & \textbf{0.261}
                     & \textbf{5.012} & \textbf{0.782} & \textbf{0.249}
                     & \textbf{5.742} & \textbf{0.731} & \textbf{0.224}\\
    \bottomrule
  \end{tabular}
\end{table*}

\subsection{Ablation Studies}
\subsubsection{Semantic Contribution Modeling}
We assess semantic--rhythm control using overall BEAT performance and a paired transcript-masking comparison. In the latter, each configuration receives the same speech audio with either the original or a masked transcript. As shown in Table~\ref{semantics}, overall SRGR increases from 0.241 with direct fusion to 0.248 with the dual branch and 0.254 with CIG. The full model reaches 0.261 and achieves the best FGD, BC, and Diversity. The CIG-only and w/o $\mathcal{L}_{\mathrm{IG}}$ rows report identical overall scores; paired masking results are additionally available for the latter. On the paired evaluation, the full model's semantic SRGR decreases from 0.260 to 0.227 after masking, a drop of 0.033, compared with a drop of 0.019 (0.251 to 0.232) without $\mathcal{L}_{\mathrm{IG}}$. The preference for gestures generated with the original transcript also increases from 62\% to 74\%. These results indicate stronger sensitivity to textual semantics under the full objective. BC decreases from 0.803 to 0.776 for the full model and from 0.796 to 0.781 without $\mathcal{L}_{\mathrm{IG}}$, showing that transcript masking also affects beat consistency.

\subsubsection{Acoustic Reliability Modeling}
We further evaluate noise-conditioned adaptive layer normalization (AdaLN) and stochastic latent perturbation by adding background noise to the BEAT test audio at 10 dB and 0 dB SNR. The variant without reliability modeling removes both components; the other variants retain AdaLN, perturbation, or their combination. As shown in Table~\ref{noise}, the full model achieves an FGD of 4.362, a BC of 0.805, and an SRGR of 0.261 on clean speech, consistent with the main results. Performance deteriorates as noise increases, particularly without reliability modeling. Both components independently improve robustness, with AdaLN yielding slightly larger gains at both noise levels. Their combination achieves the best results throughout. At 0 dB, the full model reduces FGD from 6.736 to 5.742 and increases BC and SRGR from 0.672 and 0.194 to 0.731 and 0.224, respectively. These results suggest that feature recalibration and stochastic latent perturbation provide complementary protection against corrupted acoustic inputs.

\section{Conclusion}
We proposed a reliability-aware semantic contribution modeling framework for co-speech gesture generation. The method measures text-induced predictive changes through a dual-branch conditional information-gain proxy and combines them with semantic relevance to regulate textual influence. It further adaptively balances semantic expressiveness and rhythmic synchronization to generate gestures that are both meaningful and temporally aligned with speech. To improve robustness under corrupted acoustic conditions, we introduce acoustic reliability modulation together with stochastic latent perturbation, which suppresses unreliable audio variations and stabilizes motion generation. Extensive experiments on BEAT and TED-Expressive demonstrate consistent improvements in motion quality, semantic relevance, and generation diversity.

\FloatBarrier

\end{document}